\documentclass[12pt]{article}          
\usepackage{amsmath}
\usepackage{algorithm}
\usepackage{algorithmic}
\usepackage{fancyhdr}
\usepackage{graphicx}
\begin{document}

\title{Predicting online user behaviour using deep learning algorithms}


\author{Armando Vieira        
Redzebra Analytics\\
 1 Quality Court
WC2A 1HR London, UK\\
\texttt{armando@redzebra-analytics.com}
}

\maketitle

\begin{abstract}
We propose a robust classifier to predict buying intentions based on user behaviour within a large e-commerce website. In this work we compare traditional machine learning techniques with the most advanced deep learning approaches.
We show that both Deep Belief Networks and Stacked Denoising auto-Encoders achieved a substantial improvement by extracting features from high dimensional data during the pre-train phase. They prove also to be more convenient to deal with severe class imbalance.

\smallskip
\noindent \bf{Artificial Intelligence, Auto-encoders, Deep Belief Networks, Deep Learning, e-commerce, optimisation}

\end{abstract}

\section{Introduction} \label{intro}
Predicting user intentionality towards a certain product, or category, based on interactions within a website is crucial for e-commerce sites and ad display networks, especially for retargeting. 
By keeping track of the search patterns of the consumers, online merchants can have a better understanding of their behaviours and intentions~\cite{ref5}.

In mobile e-commerce a rich set of data is available and potential consumers search for product information before
making purchasing decisions, thus reflecting consumers purchase intentions. 
Users show different search patterns, i.e, time spent per item, search frequency and returning visits~\cite{ref2}.

Clickstream data can be used to quantify search behavior using machine learning  techniques~\cite{ref6}, mostly focused
on purchase records. While purchasing indicates consumers final preferences in the same category, search is also 
an essential component to measure intentionality towards a specific category. 

We will use a probabilistic generative process to model user
exploratory and purchase history, in which the latent context variable is introduced to capture
the simultaneous influence from both time and location.  By identifying the search
patterns of the consumers, we can predict their click decisions in specific contexts
and recommend the right products.

 Modern search engines use machine learning approaches to predict user activity within web content. Popular models include logistic regression (LR) 
and boosted decision trees. 
Neural Networks have the advantage over LR because they are able to capture non-linear relationship between the input features and their "deeper" architecture has inherently greater modelling strength. On the other hand decision trees - albeit popular in this domain - face additional challenges with with high-dimensional and sparse data~\cite{ref4}.


The advantage of probabilistic generative models inspired by deep neural networks is that they can mimic the process of a consumer's purchase behaviour and capture the latent variables to explain the data.

The goal in this paper is to identify activity patterns of certain users that lead to buy sessions and then extrapolate as templates to predict high probability of purchase in related websites.
The data used consists of about 1 million sessions containing the click data of users - however, only 3\% of the training data consist of buy sessions - so making it a very unbalanced dataset. 

The rest of this paper is organized as follows: Section 2 describes the data used in our study and pre-processing methods and Non-negative Matrix Factorization for dimensionality reduction. Section 3 presents the classification algorithms. Section 4 describes in detail the deep learning algorithms (Deep Belief Networks and Stacked Denoising Auto-encoders)  and Section 5 presents the results.

\section{Data Description}
\label{sec:data}

Data consists of six months of records of user interaction with an e-commerce website. Events have a user$_{id}$, a timestamp, and event type. There are 5 categories of events: pageview of a product, basketview, buy, adclick and adview. There are around 25 000 different types of products. In case of a buy or a basketview we have information about the price and extra details. 
We ignore adview and adclick events as they are not relevant for the present propose.

The data is very sparse and high dimensional. There are two obvious ways to reduce the dimensionality of the data: 
either by marginalizing the time (aggregate pageviews per user over the period) or the product pageviews (aggregate products viewed per time frame). In this work we follow the first approach as most shopping ($\sim$ 87\%) occurs within 5 days of first visit. 

The training data is composed of a set of sessions $s \in S$ and each session contains a set of items $i \in I$ that were displayed to the user. The items that has been bought in session $s$ are denote by $B_s$. There are two types of sessions $S_b$ (the sessions that end in buying) and $S_{nb}$ (the sessions that do not end in a transaction).

Given the set of sessions $S_t$, the task is to find all the sessions $S_{b}$ which have at least one buy event. If a session $s$ will contains a buy event, we want to predict the items $B_s$ bought. Therefore we have two broad objectives: 1) classification and 2) order prediction. In this work we will focus only on first task.

The data is highly unbalanced for the two classes considered (buy and non-buy), so we face a serious class imbalance problem. Furthermore, only about 1\% of products (around 250) have a full category identification. However, this fraction corresponds to about 85\% of pageviews and 92\% of buys - so we have a very skewed distribution. 
Initially we consider only interactions with this subset of products.
The data is about 10Gb and cannot be loaded into memory, so we first took a subsample of the first 100 000 events just to have a snapshot of the interactions. We found:
\begin{itemize}
\item	78 360 pageviews events (~78.4\% of total events) from 13342 unique users. 
\item	16 409 basketview (~16.4\%) from 3091 unique users.
\item	2 430 sales events (~2.5\%) from 2014 unique users (around 1.2 sales per user). 
\end{itemize}
If we restrict to the 257 label product categories, we found 39561 pageviews, from 7469 distinct users, which is about half of the population. In this work we didn't consider time as data is very sparse and we aggregate it at several temporal basis (see Table~\ref{tabdata})


\subsection{Data preprocessing} 
Each session an unique id a timestamp is recorded for each activity in the website, so that we order users clicks on the items in a session. 
The duration of a click could easily be found by simply subtracting time of that click from the time of the next click. 
Now, for each distinct item in a session if we sum the duration of the clicks in which the item appears, we define the duration of the item in that session. 
After sorting by timestamp we append itemDuration (the time an item is inspected in a session) to each click data. 
We extract other properties, which are specific to an item and append it to each click data - see Table ~\ref{tabclick}.
We build a click-buy ratio of users by averaging the click-buy ratio of all the items in a session. 

We also used the description of the item bought, in a form of a small text. 
To handle textual data we convert words of descriptions into a 50 dimension vector using word2vec~\cite{mikolov} and used the arithmetic average of the vectors.

\begin{table}
\caption{Constructed parameters based on clickstream data }
\begin{tabular}{ll}
\hline
Symbol & Description \\
\hline
$D_s$  & Duration session before purchase \\
$C/B$  & Click to buy ratio for users\\
$S_B$ & Median number of  sessions before buy\\
$Desc$ & Description\\
$Price$ & Price of an item\\
$Duration$ & The total time spent on an item over all the sessions\\
$Hour$ & hour of the day when the session occurred\\
$Nc$ &number of clicks in a session\\
$Price$ & average items price of purchase in a session\\
$Views_{24h}$ &Number of page views in the last 24 hours\\
$Views_{week}$ & Number of page views in the last week\\
\hline
\end{tabular}
\label{tabclick}
\end{table}

To build the data set we first restrict to the set of 257 product categories. Data was aggregated at the week level per product category and semi-week (two time buckets). In this first iteration we will not add "basket view" events as most of them are made on the same session/day of sales events and the objective is to predict sales with at least one day of delay. We will consider this in next iteration.
Users with less then 10 clicks in the website were removed.
All data sets were balanced: same number of sales events and non-sales events. Due to the large size of data, we essentially study the importance of sample size and the efficiency of the algorithms dealing with the dimensionality of the the data. 
Since we want to predict purchases within a time windows of 24h, we excluded events in this period.
Next table describe the various tests done with the 6 datasets consider. The size refers to the number of buying session. All datasets were balanced by subsampling the non-buying session data.

\begin{table}
\caption{Different datasets used for testing the models}
\begin{tabular}{lll}
\hline\noalign{\smallskip}
Data1  & Size & Description \\
\noalign{\smallskip}\hline\noalign{\smallskip}
Dataset 1 & 3 000 & Sales weekly aggregated \\
Dataset 2 & 10 000 & Same as 1 but more data \\
Dataset 3 & 30 000 & Same as 1 but more data \\
Dataset 4 & 10 000 & Same as 2 but semi-weekly aggregated \\
Dataset 5 & 10 000 & Same as 1 with 2000 categories \\
Dataset 6 & 30 000 & Same as 3 with 2000 categories \\
\noalign{\smallskip}\hline
\end{tabular}
\label{tabdata}
\end{table}

Data was provided in JSON format and we sort all the click and buy sessions by session$_{Id}$. The number of sessions in our own test data was 1506453. We kept 54510 buy sessions in our test data and according to scoring.

The click data of a buy session contain a set of items bought ($Bs$).
For each item $i \in  B_s$ we extract both session-based and item-based features. 

\subsection{Non-Negative Matrix Factorization}
In order to test the impact of excluding some product categories we consider Data 5 with the top 2000 more visited product categories. 
Since this a huge dimensional search space, we used Non-Negative Matrix Factorization (NMF) to reduce the dimensionality. NMF is a class of unsupervised learning algorithms~\cite{refnmf}, such as Principal Components Analysis (PCA) or learning vector quantization (LVQ) that factorizes a data matrix subjected to constraints. Although PCA is a widely used algorithm it has some drawbacks, like its linearity and poor performance on factors. Furthermore, it enforces a weak orthogonality constraint. LVQ uses a winner-take-all constraint that results in clustering the data into mutually exclusive prototypes but it performs poorly on high dimensional correlated data. Given a non-negative matrix $V$ (containing the training data), NMF learns non-negative matrix factors, $W$ and $H$, such that: $V\cong WH$

Each data vector $V$ (data entry) can be approximated by a linear combination of the columns of $W$, weighted by the patterns matrix $H$. Therefore, $W$ can be regarded as containing a basis for the linear approximation of the data in $V$. Since relatively few basis vectors are used to represent many data vectors, good approximation can only be achieve if the basis vectors discover the structure that is latent in the data.

NMF was successfully applied to high dimensional problems with sparse data, like image recognition and text analysis. In our case we used NMF to compress data into a feature subset. 
The major issue with NMF is the lack of an optimal method to compute the factor matrixes and stopping criteria to find the ideal number of features to be selected. 


\section{Classifiers}
Our task is divided in to two subtasks: i) predicting the outcome of a session and ii) predict the set of items that should be bought in that session. 
Two set of classifiers are involved: binary and ranking prediction.
Building a single classifier is not advisable due to the large dimensionality of the problem.

Based on the data sets, we test the performance of two classifiers: Logistic Regression and Random Forest. The first is a standard in industry and serve as a baseline the second is more robust and produce in general better results. It has the disadvantage of their predictions not being ease to understand (black box). We used the algorithms without any optimization of the parameters (number of trees, numbers of variables to consider in each split, split level, etc.)
As a KPI to measure performance we use the standard Area Under Roc curve (AUC). An AUC=0.5 meaning a random (useless) classifier and 1 a perfect one. For all runs we used 10 fold cross validation. 

\subsection{Decision Trees}
Decision trees possess several inherent advantages over other classification methods such as support vector machines, neural networks, linear regression and logistic regression. Decision trees are:
\begin{itemize}
\item  Extremely easy to visualize and interpret: a decision tree can be represented graphically, allowing the user to actually see the structure of the classifier;
\item White-box models: by observing a decision tree, one can clearly understand all the intermediate steps of the classification process, such as which variables are used, by what order, etc. This is not true for other methods such as neural networks, whose parameters cannot be directly interpreted;
\item Extremely fast: decision trees are trained in a relatively short time and are particularly fast in classifying new data.
\end{itemize}

However, decision trees possess several drawbacks. The process of building an optimal decision tree can be proved to be NP-hard, and therefore it is not possible to create a globally optimal tree. Decision trees will often overfit the data unless some regularization methods, such as pruning, or imposing a minimum number of training samples per leaf, are used. Also, because of the characteristics of the cost function used to determine the best split at a node, trees will tend to prefer categorical variables with more categories over other variables. This may cause the classifier to incorrectly consider these variables as more important than those with fewer categories.

\subsection{Random Forest}
The Random Forest (RF) algorithm creates an ensemble of decision trees using randomization. When an input is to be classified, each tree classifies the input individually. The final classification is then decided by choosing the majority vote over all the trees. The likelihood of a certain input belonging to each class is computed by averaging the probabilities at the leaves of each tree.

Each tree is grown in an independent, random way. The set that is used to train a given tree is a subset of the original training data; each training example is selected at random (with replacement) from the original data set. At each node of the tree, rather than testing the best split among all the attributes, only a randomly chosen subset of the attributes (which is usually much smaller than the full set of attributes) are used for determining the best split. Each tree is grown to its full extent, meaning that no pruning occurs.

The final classifier is efficient and capable of dealing with large data sets (i.e., data that contains a large number of variables), missing data, and outliers. In the present problem, there is a large amount of information available for each client. In order to avoid the deletion of possibly significant variables in order to reduce the data to a manageable size - something which would be mandatory if neural networks were used, for example - random forest is the algorithm of choice.

Random forest retain the strengths of decision trees  while countering some of their disadvantages. Even if the trees in the forest are grown without pruning, the fact that the classifier?s output depends on the whole set of trees and not on a single tree, the risk of overfitting is considerably reduced. The randomness that is introduced in the creation of each tree also prevents the classifier from memorizing all the examples in the training set. The regularization techniques mentioned in the previous paragraph can also be applied to the trees in the forest, further reducing the risk of overfitting. However, random forests have the same bias towards variables with many categories as decision trees.

\section{Deep Learning Methods}
Deep learning refers to a wide class of machine learning techniques and architectures, with the hallmark of using many layers of non-linear processing that are hierarchical in nature~\cite{hinton}. The concept of deep learning originated from artificial neural network research - feed-forward neural networks or MLPs with many hidden layers refereed as deep neural networks (DNNs). These networks are generally trained by a gradient descent algorithm designated Back-propagation (BP). However, for deep networks, BP alone has several problems: local optima traps in the non-convex objective function and vanish gradients (learning signal vanish exponentially as information in backpropagated through layers).

In this section we will introduce two deep learning approaches to handle the high dimensionality of the search space and compare performance with logistic regression and random forest algorithms.

\subsection{Deep Belief Networks}
In 2006 Hinton proposed an unsupervised learning algorithm for a class of deep generative models, called deep belief networks (DBN)~\cite{hinton1}. A DBN is composed of a stack of restricted Boltzmann machines (RBMs). A core component of the DBN is a greedy, layer-by-layer learning algorithm which optimizes DBN weights. Separately, initializing the weights of an MLP with a correspondingly configured DBN often produces much better results than that with the random weights. 

DBN belongs to a class of energy based models. In this case the algorithm runs as follows: 

      For a given RBM, we relate the units with the energy function,
            \begin{equation}Energy(v,h)=-b'h-c'v-h'Wv.\end{equation}
            where $b,c$ are offsets/biases and $W$ comprises the weights connecting units
      The joint probability of the visible (v) and hidden (h) unities, (v,h) is
                        \begin{equation} P(v,h)=\frac{1}{Z}e^{-Energy(v,h)}\end{equation}
            where $Z$ is the normalization term.

 We obtain the free energy form by marginalizing $h$
             \begin{equation}P(v)=\frac{\sum_h e^{-Energy(v,h)}}{Z}=\frac{e^{-FreeEnergy(v)}}{Z}\end{equation}
Taking advantage of free energy form makes it easier to compute gradients with visible units only.

We rewrite the energy function into the form,
            \begin{equation}Energy(v,h)=-\beta(v)-\sum_i\gamma_i(v,h_i).\end{equation}
Then we factorize $P(v)$
          \begin{align*}
            P(v) &=\frac{\sum_h e^{-Energy(v,h)}}{Z}=\frac{e^{-FreeEnergy(v)}}{Z}\\
                 &=\frac{1}{Z}\sum_{h_1}\sum_{h_2}\cdots\sum_{h_k}e^{\beta(v)-\sum_i\gamma_i(v,h_i)}
                  =\frac{1}{Z}\sum_{h_1}\sum_{h_2}\cdots\sum_{h_k}e^{\beta(v)}\prod_ie^{-\gamma_i(v,h_i)}\\
                 &=\frac{e^{\beta(v)}}{Z}\sum_{h_1}e^{-\gamma_1(v,h_1)}\sum_{h_2}e^{-\gamma_2(v,h_2)}
                   \cdots\sum_{h_k}e^{-\gamma_k(v,h_k)}\\
                 &=\frac{e^{\beta(v)}}{Z}\prod_i\sum_{h_i}e^{-\gamma_i(v,h_i)}\\
          \end{align*}

\begin{figure}
  \includegraphics[width=\linewidth]{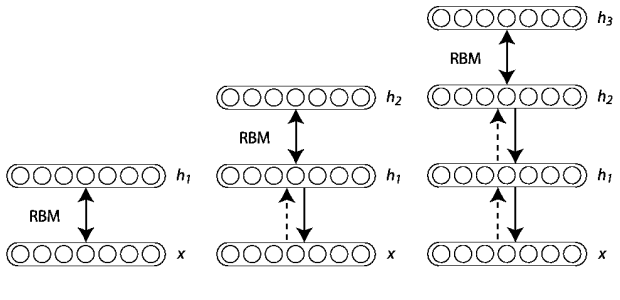}
  \caption{Structure of Deep Belief Network.}
  \label{fig:dbn}
\end{figure}

DBN were been used for a large variety of problems, ranging from image recognition, recommendation algorithms and topic modelling. 
In addition to the supply of good initialization points for a multilayer network, the DBN comes with other attractive properties: the learning algorithm makes effective use of unlabeled data; ii) it can be interpreted as a probabilistic generative model and iii) the over-fitting problem, which is often observed in the models with millions of parameters such as DBNs, can be effectively alleviated by the generative pre-training step.       
The downside of DBN is that they are hard to train and very sensitive to learning parameters like weights initialisation. 

\subsection{Auto-encoders}
Autoencoders are  a representation learning technique using unsupervised pre-training to learn good representations of the data transform and reduce the dimensionality of the problem in order to facilitate the supervised learning stage. 

An autoencoder is a neural network with a single hidden layer and where the output layer and the input layer have the same size. Suppose that the input $x \in {R}^m$ and suppose that the hidden layer has $n$ nodes. 
Then we have a weight matrix $W\in {R}^{m\times n}$ and bias vectors $b$ and $b'$ in $R^m$ and $R^n$, respectively. 

Let $s(x) = 1/(1+e^{-x})$ be the sigmoid (logistic) transfer function. Then we have a neural network as shown in Fig.~2. When using an autoencoder to encode data, we calculate the vector $y=s(Wx + b)$; corresponding when we use an autoencoder to decode and reconstruct back the original input, we calculate $z=s(W^{T}x+b^{'})$. 
The weight matrix of the decoding stage is the transpose of weight matrix of the encoding stage in order to reduce the number of parameters to learn. We want to optimize $W$, $b$, and $b^{'}$ so that the reconstruction is as similar to the original input as possible with respect to some loss function. The loss function used is the least squares loss: 
\begin{equation}
E(t,z)=\frac{1}{2}(t-z)^2
\end{equation}
where $t$ is the original input. After an autoencoder is trained, its decoding stage is discarded and the encoding stage is used to transform the training input examples as a preprocessing step.

\begin{figure}
  \includegraphics[width=\linewidth]{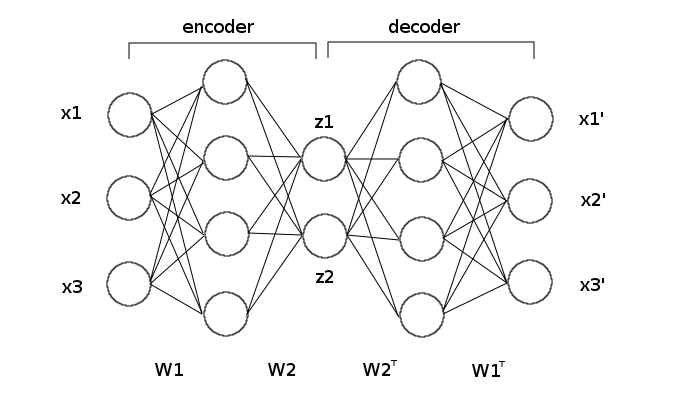}
  \caption{Structure of an autoencoder. The weights of the decoder are the transpose of the encoder.}
  \label{figautoencoder}
\end{figure}

Once an autoencoder layer has been trained, a second autoencoder can be trained using the output of the first autoencoder layer. This procedure can be repeated indefinitely and create stacked autoencoder layers of arbitrary depth. It is been shown that each subsequent trained layer learns a better representation of the output of the previous layer. Using deep neural networks such as stacked autoencoders to do representation learning is also called deep autoencoders - a subfield of machine learning. 

For ordinary autoencoders, we usually want that $n<m$ so that the learned representation of the input exists in a lower dimensional space than the input. This is done to ensure that the autoencoder does not learn a trivial identity transformation. A variant is the denoising autoencoders that uses a different reconstruction criterion to learn representations \cite{vincent2010stacked}. This is achieved by corrupting the input data and training the autoencoder to reconstruct the original uncorrupted data. By learning how to denoise, the autoencoder is forced to understand the true structure of input data and learn a good representation of it. When trained with a denoising criterion, a deep autoencoder is also a generative model.
Although the loss function $E(t,z)$ for neural networks in general is non-convex, stochastic gradient descent (SGD) is sufficient for most problems and we use it in this work. 

\subsection{Autoencoders formulation}
The derivative of the output error $E$ with respect to an output matrix weight $W_{ij}^O$ is as follows.

\begin{align}
\begin{split}
\frac{\partial E}{\partial W^O_{ij}} &= \frac{\partial E}{\partial z_j}\frac{\partial z_j}{\partial W^O_{ij}} \\
																	 &=(z_j - t_j)\frac{\partial s(n_j)}{\partial x_j}\frac{\partial x_j}{\partial W^O_{ij}} \\
																	 &=(z_j-t_j)s(n_j)(1-s(n_j))x_i \\
																	 &=(z_j-t_j)z_j(1-z_j)x_i \\
\end{split}
\label{}
\end{align}

Now that we have the gradient for the error associated to a single training example, we can compute the updates.
\begin{equation}
\begin{split}
\delta^O_j &= (z_j-t_j)z_j(1-z_j) \\
W^O_{ij} &\leftarrow W^O_{ij} - \eta \delta^O_j x_i \\
b^O_j &\leftarrow b^O_j - \eta\delta^O_j
\end{split}
\end{equation}

The computation of the gradient for the weight matrix between hidden layers is similarly easy to compute.
\begin{align}
\begin{split}
\frac{\partial E}{\partial W^H_{ij}} &= \frac{\partial E}{\partial y_j}\frac{\partial y_j}{\partial W^H_{ij}} \\
&=\left(\sum_{k=1}^m \frac{\partial E}{\partial z_k}\frac{\partial z_k}{\partial n_k}\frac{\partial n_k}{\partial y_j} \right)\frac{\partial y_j}{\partial n_j}\frac{\partial n_j}{\partial W_{ij}^H}\\
																	 &=\left(\sum_{k=1}^m (z_k - t_k)(1-z_k)z_kW_{jk}^O \right)y_j(1-y_j)x_i
\end{split}
\label{}
\end{align}

And then using the computed gradient we can define the updates to be used for the hidden layers
\begin{align}
\begin{split}
\delta^H_j &= \left(\sum_{k=1}^m (z_k - t_k)(1-z_k)z_kW_{jk}^O \right)y_j(1-y_j) \\
W^H_{ij} &\leftarrow W^H_{ij} - \eta\delta^H_jx_i \\
b^H_j &\leftarrow b^H_j - \eta\delta^H_j
\end{split}
\end{align}

In general, for a neural network we may have different output error functions and these will result in different update rules. We will also give the updates for the cross-entropy error function with softmax activation in the final layer.
The cross entropy error function is given by \[E(x,t) = -\sum_{i=1}^n \left(t_i\ln z_i + (1-t_i)\ln(1-z_i)\right)\]
and the softmax function is given by $\sigma(x_j) = e^{x_j} /(\sum_k e^{x_k})$. Following the same procedure as above
for computing the gradient and the updates, we find that for hidden/output layer

\begin{equation}
\begin{split}
\frac{\partial E}{\partial W^O_{ij}} &= (z_j - t_j)y_i \\
\delta^O_j &= (z_j-t_j) \\
W^O_{ij} &\leftarrow W^O_{ij} - \eta \delta^O_j x_i \\
b^O_j &\leftarrow b^O_j - \eta\delta^O_j.
\end{split}
\end{equation}

We find that the updates for the hidden layer is the same as in the squared error loss function with sigmoid activation.
\begin{algorithm}
\label{bp}
\begin{algorithmic}
\FOR {$t=T,\ldots,1$}  
\STATE to compute $\frac{\partial E}{\partial net_t}$, inititalize real-valued error signal variable $\delta_t$ by 0;
\STATE if $x_t$ is an input event then continue with next iteration;
\STATE if there is an error $e_t$ then $\delta_t := x_t-d_t$;
\STATE add to $\delta_t$ the value $\sum_{k \in out_t}  w_{v(t,k)} \delta_k$;
\STATE multiply  $\delta_t$ by $f'_t(net_t)$;
\STATE for all $k \in in_t$ add to  $\triangle_{w_{v(k,t)}}$ the value $x_k \delta_t$
\ENDFOR
\STATE change each $w_i$ in proportion to $\triangle_i$ and a small real-valued learning rate
\end{algorithmic}
\caption{Algorithm for Auto Encoders}
\end{algorithm}

The algorithm and derivations for the auto-encoder are a slight variation on the above derivations for a more general neural network. The weight matrix of the output layer (decoding stage) is the transpose of the weight matrix of the hidden layer (encoding stage). Thus $z=s(W^{O}(W^{H}x + b) + b^{'})$, $(W^H)^T = W^O$, and $W^H_{ij} = W^O_{ji}$. For training denoising autoencoders in particular, $z=s(W^{O}(W^{H}x_{\text{corr}} + b) + b^{'})$, where $x_{\text{corr}}$ is a randomly corrupted version of the original input data $x_{\text{orig}}$ and the loss function is defined as $E(x_{\text{orig}}, z)$. In order words, we are trying to learn an autoencoder takes in corrupted input and reconstructs the original uncorrupted version. Once we have trained a single autoencoder layer, we can stack another autoencoder layer on top of the first one for further training. This second autoencoder takes the corrupted output of the hidden layer (encoding stage) of the first autoencoder as input and is again trained to minimize the loss function.

\subsection{Regularization}
Avoiding overfiting is especially crucial for deep neural nets with typically have millions of parameters. 
DBN can generate large and expressive models capable of representing complex dependencies between inputs and outputs. Generative unsupervised pre-training~\cite{hinton_1} is a powerful data-dependent regularizer, while dropout is the most commonly used.

L2 regularization shifts the weights towards zero which may not be desirable.
Dropout penalizes large weights that result in uncertain predictions or hidden unit activations. Another way to view dropout is as approximate model averaging over the exponentially numerous different neural nets produced pruning random subsets of
hidden units and inputs. 
In this work we used dropout regularization.

\section{Results}
First we run the algorithms using only the aggregated variables from Table~\ref{tabclick} to predict buying events. Since this data is low dimensional, we only consider LR and RF algorithms. Note, however, that most buying events occur within the 24h time-frame. 
We found and AUC for LR of 0.58 and for RF of 0.61.

Then we used all data combining Table 1 with Table 2 and the description about products using a 50 dimension vector composition (excluding a set of stop words). Results are present in Table~\ref{tab:results1}

\begin{table}[h]
\centering
\begin{tabular}{l | l l l l}
Data set & LR& RF  \\  \hline                                                        
Data 1	&0.67&0.71\\
Data 2&	0.69	&0.76\\
Data 3&	0.70	&0.80\\
Data 4&	0.68	&0.82\\
Data 5-100&	0.62	&0.67\\
Data 5-200&	0.64	&0.69\\
Data 5-300&	0.64	&0.72\\
\end{tabular}
\caption{Results for AUC with Random Forest and Logistic Regression.}
\label{tab:results1}
\end{table}

We conclude that sample size is an important factor in the performance of the classifier, though the Logistic Regression does not have the same gains as the Random Forest (RF) algorithm. 

From data set 4 we also conclude that time of events is an important factor to take into account: although we increase the dimensionality of the search space, we still have a net gain even using fewer training examples.

From data set 5, we concluded that the NFM algorithm is doing some compression on data but not in a very efficient way (only the data with 300 features had improved the accuracy over the initial subset of products). In next section we suggest using Auto-encoders to reduce the dimensionality of data for all the 25 000 categories.

Quite surprisingly, we found that the use of detailed information about which products the user visited does not carry much gain to the logistic regression accuracy (in some cases it even decreases - probably due to the increase of dimensionality), while RF can capture higher accuracies.


\subsection{Deep Learning results}
One of the main advantages of DBN or SdA is that we can use all the available data (even if unlabeled) to pre-train the model in an unsupervised, or generative, way. In DBN this is intended to capture high-order correlation of the observed or visible data for pattern analysis or synthesis purposes when no information about target class labels is available. Then, we can jointly characterize statistical distributions of the visible data and their associated classes, when available. Finally the use of Bayes rule can turn this type of generative networks into discriminative machines.

We used all one million session data (pageviews of products aggregated per user and per week) together with the composed parameters described in Table~\ref{tabclick}.

For each assay, we held out at random 25\% of data to use as a test set, leaving the remaining
75\% as a training set. We split the training set into four folds and trained each model four times with
a different fold held out as validation data. We average the test set AUCs of the four models when
reporting test set results. We used performance on the validation data to select the best particular
model in each family of models. To the extent that the baseline models required metaparameter tuning
(e.g. selecting the number of trees in the ensemble), we performed that tuning by hand using validation
performance.

Neural networks have many metaparameters: architectural, such as layer sizes and hidden unit transfer functions; optimization, such as learning rates and momentum values; and regularization, such as the dropout probabilities for each layer.
Deep Neural Networks can have a very large number of parameters, in our case, between one and 4 million weights. 
All neural net metaparameters were set using Bayesian optimization to maximize the validation AUC. Bayesian optimization is ideally suited for globally optimizing blackbox, noisy functions while being parsimonious in the number of function evaluations.

Deep Neural networks require careful tuning of numerous metaparameters, which is one of the hardest and time consuming tasks in implementing  these solutions. We have to go thorough exploration of architectures and metaparameters such as regularisation
parameters and weights initialisation. We used the constrained version of Spearmint of Snoek et al. ~\cite{snoek}
with warping enabled and labeled training runs that diverged as constraint violations. We let Spearmint optimize the metaparameters listed below with a budget of 20 trials. The ranges were picked based on iteration on the first single hidden layer. 

The metaparameters considered to train the networks were:

\begin{itemize}
\item dropout fraction $\in$ [0: 0.3]
\item number of training epochs $\in$ [10: 100] for nets with a single hidden layer and $\in$ [10: 150] for nets with two or more hidden layers
\item number of hidden units in each layer. No hidden layer was allowed more than 500 units. The minimum number of hidden units in a layer for a single task neural net was 16 in our first single hidden layer and 64 all other times.
\item the annealing delay fraction $\in$ [0: 1] is the fraction of the training iterations that
must complete before we start annealing the learning rate
\item  The initial  learning rate applied to the average gradient over a minibatch $\in$ [0.001: 0:25] 
\item momentum $\in$ [0: 0.95]
\item the L2 weight cost  $\in$ [0: 0.01] 
\item the hidden unit activation function, either logistic sigmoids or rectified linear units - all hidden units in a network use the same activation function.
\item the noise level applied to the input layer (only for the SdA) $\in$ [0: 0.2].

\end{itemize}
  
To run the network models we used the implementation Keras based on theano libraries (see http://keras.io/). 
A softmax layer was attached to the last hidden layer of the pre-trained network for the supervision phase. 
We optimize the metaparameters of the networks on dataset 3 (30 000 purchase transactions) and used the same parameters on other datasets. 
We found that dropout for ReLU nets are always kept while for sigmoid transfer functions it was rarely greater then zero.
For the unsupervised we used all data available.

The results are presented in Table~\ref{tabresults2}. We can see that networks pre-trained with Stacked denoising Autoencoders 
reach the highest accuracy, which may be due to the fact that we have a very sparse data set. Improvements as compared with other traditional methods are notorious.

\begin{table}[h]
  \centering
\begin{tabular}{l | l l}
Data set & DBN&SdA \\ \hline                                                        
Data 3&	0.82	&0.83\\
Data 6&	0.84	&\textbf{0.86}\\

\end{tabular}
\caption{Results for AUC for classification of purchase likelihood using DBN and SdA.}
\label{tabresults2}
\end{table}



\begin{figure}
  \includegraphics[width=\linewidth]{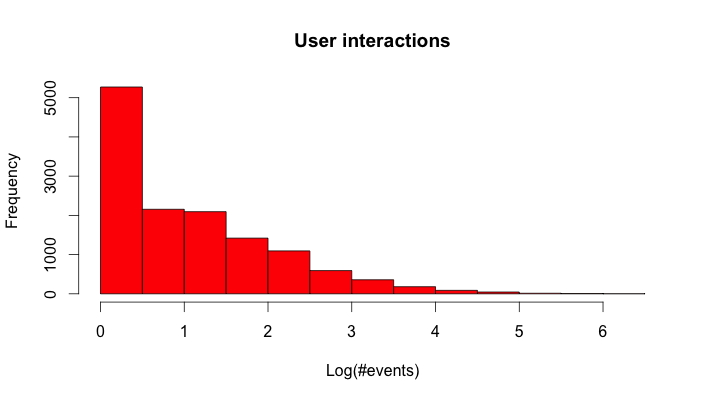}
  \caption{Distribution of pageview events for a subset of the data.}
  \label{fig:test}
\end{figure}



\section{Conclusions and future work}
In this paper we apply different machine learning algorithm, including deep neural networks for modelling purchase prediction.
We showed that using boosting methods like random forest improves performance over linear model
like logistic regression.

To our knowledge this is the first time that deep architectures like DBN and DAE were applied to e-commerce platform for modelling user behaviour.
Finally we did a comparison and stated improvements of using deep neural networks with existing algorithms.

Further research can include testing on real-time data, and see the performance effects on a real-time.
However, more work would need to be done on improving time efficiency of the
In terms of scalability, the data is extremely sparse and the class of algorithms we used does not parallelize well with multiple cores. As the results clearly show gap in performance improving with large data size, it would be interesting
to see the effect of using much larger training data. Moreover, since many of the ID-based features
are in forms of words it may be useful to initialize the neural network as an RBM trained with
unsupervised contrastive divergence on a large volume of unlabaled examples. And then fine tune
it as a discriminative model with back propagation. It could also prove useful to train multiple
networks in parallel and feed all of their outputs individually to MatrixNet, as feature vectors, instead
of just a single average.


\bibliographystyle{spr-chicago}      

\end{document}